\documentclass[11pt]{article}

\usepackage[preprint]{acl}

\usepackage{times}
\usepackage{latexsym}
\usepackage[T1]{fontenc}
\usepackage[utf8]{inputenc}
\usepackage{microtype}
\usepackage{inconsolata}
\usepackage{graphicx}
\usepackage{pdflscape}

\usepackage{booktabs}
\usepackage[most]{tcolorbox}
\usepackage{amsmath}
\usepackage{multirow}
\usepackage{url}
\usepackage{comment}
\usepackage{amssymb}
\usepackage{siunitx}
\usepackage{subcaption}
\usepackage{tikz}
\usetikzlibrary{arrows.meta, positioning, calc, fit}
\usepackage{float}
\usepackage{array}
\usepackage{verbatim}

\usepackage{xcolor}
\usepackage{xspace}

\definecolor{percolor}{RGB}{220,20,60}     
\definecolor{orgcolor}{RGB}{0,102,204}     
\definecolor{loccolor}{RGB}{0,153,51}      
\definecolor{datecolor}{RGB}{153,51,255}   

\usepackage{soul,xcolor}

\sethlcolor{percolor!15}
\newcommand{\PER}[1]{{\sethlcolor{percolor!15}\hl{\textbf{#1}}}$_{\textsc{PER}}$}
\newcommand{\ORG}[1]{{\sethlcolor{orgcolor!15}\hl{\textbf{#1}}}$_{\textsc{ORG}}$}
\newcommand{\LOC}[1]{{\sethlcolor{loccolor!15}\hl{\textbf{#1}}}$_{\textsc{LOC}}$}
\newcommand{\DATE}[1]{{\sethlcolor{datecolor!15}\hl{\textbf{#1}}}$_{\textsc{DATE}}$}

\newtcolorbox{example}[1]{
    float=t,
    floatplacement=t,
    colback=gray!3,
    colframe=black!60,
    title={#1},
    boxrule=0.5pt,
    arc=2mm,
    before upper={\raggedright}
}

\title{RunyaNER: Auxiliary Language Selection for Runyankore NER}

\author{
Prosper Arineitwe Asiimwe \hspace{1em}
Francois Meyer \hspace{1em}
Jan Buys \\
Department of Computer Science, University of Cape Town \\
\texttt{arnari002@myuct.ac.za, \{francois.meyer, jan.buys\}@uct.ac.za}
}

\begin{document}
\maketitle

\begin{abstract}
Cross-lingual zero-shot transfer and multilingual fine-tuning are promising approaches for NLP tasks such as Named Entity Recognition (NER) in low-resource languages, but in the absence of target language benchmarks, it is unclear which auxiliary language selection strategy leads to the best transfer. We introduce RunyaNER,\footnote{Dataset available at \href{https://huggingface.co/datasets/uctnlp/runyaner}{\tt huggingface.co/uctnlp/runyaner}.} the first publicly available NER benchmark for the East African language Runyankore, and use it to investigate the choice of which languages to use for transfer.  
Created with a semi-automated pipeline and fully manually verified, RunyaNER contains over 237k annotated words across 30k sentences. 
We benchmark pretrained models on RunyaNER, establishing that our dataset is of sufficient quality and size to produce effective Runyankore NER models. 
We then use RunyaNER to investigate auxiliary language selection in cross-lingual zero-shot and multilingual fine-tuning settings. 
Our experiments show that while transfer performance is highly sensitive to auxiliary language selection, embedding-based measures computed from labelled training spans correlate more strongly with downstream transfer performance than traditional linguistic features based on metadata or typology. 
By releasing RunyaNER and providing a systematic analysis of auxiliary language selection strategies, this work contributes both a new benchmark resource and practical insights for multilingual transfer in low-resource settings.

\end{abstract}

\section{Introduction}

Despite rapid advances in NLP, many African languages remain severely underrepresented in computational resources, benchmark datasets, and pretrained language models (PLMs) \citep{mussandi-wichert-2024-nlp}. This imbalance limits the development of downstream NLP systems for large populations of speakers across the continent. In particular, many African languages still lack the annotated datasets required to train and evaluate PLMs on structured prediction tasks such as NER.

Runyankore, a Bantu language spoken by over two million people in western Uganda \citep{Namyalo2016, ethnologue2025}, is one such underrepresented language. It belongs to the Runyakitara macrolanguage, which also includes closely related varieties such as Rukiga, Runyoro, and Rutooro \citep{Bernsten_1998}. Although some efforts have begun to develop computational resources for Runyankore and related languages \citep{Katushemererwe_2020, Bamutura_2020}, publicly available benchmark datasets for downstream NLP tasks remain scarce. 

\begin{table}[t]
\centering
\small
\begin{tabular}{lrrr}
\toprule
\textbf{Split} & \textbf{Sentences} & \textbf{Tokens} & \textbf{Named entity spans} \\
\midrule
Train & 15,001 & 118,622 & 2,720 \\
Dev   & 7,498  & 59,028  & 1,364 \\
Test  & 7,508  & 59,426  & 1,427 \\
\midrule
Total & 30,007 & 237,076 & 5,511 \\
\bottomrule
\end{tabular}
\caption{Summary statistics for RunyaNER.}
\label{tab:runyaner_stats}
\vspace{-0.5cm}
\end{table}

Beyond the absence of resources, an additional challenge for low-resource NLP is determining how to effectively leverage supervision from related languages for which annotated datasets are available. Multilingual PLMs \citep{devlin-etal-2019-bert, conneau-etal-2020-unsupervised} enable cross-lingual transfer through shared multilingual representations, but transfer performance varies across languages and tasks \citep{lauscher-etal-2020-zero, xu-etal-2022-cross}. Recent work has emphasised the influence of auxiliary language selection, particularly in zero-shot settings where no target-language supervision is available \citep{adelani-etal-2022-masakhaner, ERONEN2023103250}. However, it remains unclear which notions of language similarity provide the most reliable guidance for selecting effective transfer languages.

In this work, we address both challenges by introducing RunyaNER, the first publicly available NER dataset for Runyankore, and using it to systematically study auxiliary language selection for low-resource transfer. RunyaNER consists of 237,076 annotated words (5,511 named entity spans) across 30,007 sentences. The dataset was created by annotating publicly available corpora using a semi-automated human-in-the-loop workflow, with all annotation and verification performed by one of the authors, an L1 speaker of Runyankore. 

We establish the first benchmark for Runyankore NER under monolingual fine-tuning, cross-lingual zero-shot transfer, and multilingual fine-tuning settings. We further investigate auxiliary language selection across 20 African languages from MasakhaNER~2.0 \citep{adelani-etal-2022-masakhaner}, comparing metadata-based, typological, and embedding-based similarity measures. Our goal is to determine whether embedding-based similarity measures provide better guidance for auxiliary language selection than traditional linguistic resources. 

Under cross-lingual zero-shot transfer, performance is highly sensitive to auxiliary language choice, with closely related Great Lakes Bantu languages yielding the strongest transfer. However, across the full set of auxiliary languages, embedding-based similarity measures computed from labelled training spans, particularly prototype cosine similarity and Sliced Wasserstein Distance (SWD), correlate more strongly with downstream transfer effectiveness than metadata-based or typological similarity measures. In contrast, once modest amounts of Runyankore supervision are introduced through multilingual fine-tuning, differences between auxiliary language choices become substantially smaller.

To our knowledge, this is the first systematic study of auxiliary language selection for African NER using both linguistic resource and embedding-based similarity measures. 
In addition to introducing the first publicly available NER dataset for Runyankore, our findings provide practical guidance for multilingual transfer in low-resource settings and highlight the value of embedding-based similarity, when a labelled target training split is available for selecting effective auxiliary languages under weak NER supervision. 

\section{Background}

\subsection{NER with Multilingual PLMs}

Multilingual PLMs such as mBERT \citep{devlin-etal-2019-bert} and XLM-R \citep{conneau-etal-2020-unsupervised} enable cross-lingual transfer by learning shared representations across languages. This allows supervision from high-resource languages to benefit low-resource languages in tasks such as NER \citep{pires-etal-2019-multilingual, lauscher-etal-2020-zero}, which has improved NER performance for many African languages \citep{adelani-etal-2022-masakhaner}. 

However, transfer performance varies substantially across languages and depends on factors such as tokenisation coverage \citep{rust-etal-2021-good}, morphological and syntactic divergence \citep{lin-etal-2019-choosing}, and the degree of language representation in multilingual pretraining corpora \citep{conneau-etal-2020-unsupervised, ogueji-etal-2021-small}. Prior work has also shown that transfer is generally more effective between linguistically related languages \citep{lauscher-etal-2020-zero, sainik-etal-2023-transfer}. Regionally focussed models such as AfriBERTa \citep{ogueji-etal-2021-small} and Afro-XLMR \citep{alabi-etal-2022-adapting} further demonstrate the importance of language relatedness in multilingual transfer for African NLP.

\subsection{Auxiliary Language Selection for Cross-Lingual Transfer}

A key challenge in cross-lingual learning is selecting auxiliary languages that maximise transfer performance. Existing approaches derive similarity metrics either from external linguistic resources or from multilingual representation spaces. Linguistic similarity measures are typically based on resources such as URIEL/lang2vec \citep{littell-etal-2017-uriel} and LinguaMeta \citep{ritchie-etal-2024-linguameta-unified}, which capture genealogical, phonological, syntactic, and geographic properties. Embedding-based approaches instead measure similarity directly in multilingual embedding spaces, including cosine similarity between mean-pooled embeddings \citep{artetxe-schwenk-2019-massively} and distribution-based metrics such as Sliced Wasserstein Distance (SWD) \citep{nguyen2020distributionalslicedwassersteinapplicationsgenerative}. 

Several studies have shown that embedding-based similarity metrics correlate more strongly with downstream transfer performance than manually engineered typological or metadata-based features \citep{shaffer-2021-language-clustering, philippy-etal-2023-identifying, yu-etal-2021-language}. However, findings vary across tasks, models, and language families, making it unclear which similarity measures provide the most reliable guidance for realistic low-resource settings such as African-language NER.

\begin{tcolorbox}[
colback=white,
colframe=black!50,
boxrule=0.4pt,
arc=1.5pt,
left=4pt,
right=4pt,
top=4pt,
bottom=4pt,
]

{\small\bfseries Example 1}

\vspace{0.2em}

{\scriptsize\textsc{Runyankore}}
\\
\PER{Elly Karuhanga} akahwihwisa ahabwa \PER{Kadaga} kutaayaayira faamu ya \PER{Museveni}.

\vspace{0.25em}

{\scriptsize\textsc{English}}
\\
\textbf{Elly Karuhanga} murmurs over \textbf{Kadaga}'s trip to \textbf{Museveni}'s ranch.

\vspace{0.45em}
\hrule
\vspace{0.45em}

{\small\bfseries Example 2}

\vspace{0.2em}

{\scriptsize\textsc{Runyankore}}
\\
\ORG{African Union} omuri \LOC{Addis Ababa} \DATE{omukwezi kw'okubanza 2017}.

\vspace{0.25em}

{\scriptsize\textsc{English}}
\\
\textbf{African Union} in \textbf{Addis Ababa} in \textbf{January 2017}.

\end{tcolorbox}

\section{RunyaNER Dataset}

RunyaNER is constructed using two publicly available Runyankore-English parallel corpora: the Sunbird African Language Technology (SALT) corpus\footnote{\url{https://github.com/SunbirdAI/salt-data-archive}} and the Multilingual Parallel Text Corpora (MPTC) \citep{DVN_BEROE0_2023}. SALT provides broader domain coverage, including news, public communication, and conversational content, while MPTC contributes shorter and structurally more regular sentences. After preprocessing and filtering, the combined corpus contains 30,007 annotated Runyankore sentences, of which approximately 72\% originate from SALT and 28\% from MPTC.

All sentences were annotated following the MasakhaNER~2.0 guidelines \citep{adelani-etal-2022-masakhaner} using four entity types: \textsc{PER} (person), \textsc{LOC} (location), \textsc{ORG} (organisation), and \textsc{DATE} (date). All annotation and verification were performed by one of the authors, an L1 speaker of Runyankore, and no external annotators were recruited. Before labelling, the annotator studied the MasakhaNER~2.0 guidelines and applied them to a short practice sample. Quality is therefore operationalised as exhaustive guideline-based correction by a single trained L1 speaker rather than inter-annotator agreement. To reduce annotation time, we adopted a semi-automated human-in-the-loop workflow. An initial seed set of approximately 500 sentences, drawn from both source corpora and including both entity-bearing and entity-free examples, was annotated manually. We then fine-tuned XLM-R \citep{conneau-etal-2020-unsupervised} on this seed data together with Luganda training data from MasakhaNER~2.0, and used the resulting model to pre-annotate the remaining corpus. All predicted labels were reviewed and corrected manually in Doccano.\footnote{\url{https://github.com/doccano/doccano}} Sunbird's Sunflower machine-translation system\footnote{\url{https://sunflower.sunbird.ai/}} was consulted only when the English parallel was missing or fragmentary, and only as a reading aid.

Before annotation, both source corpora were normalised to remove encoding artefacts, irregular spacing, and malformed punctuation. Sentences containing severe formatting or structural errors were discarded. The final dataset was split into train, development, and test partitions using a stratified procedure to preserve consistent entity distributions across splits. Table~\ref{tab:runyaner_stats} summarises the dataset statistics, while Examples~1 and~2 illustrate annotated Runyankore sentences with their corresponding entity labels.

\begin{table*}[t!]
\centering
\scriptsize
\resizebox{\textwidth}{!}{
\begin{tabular}{l *{15}{c}}
\toprule
& \multicolumn{3}{c}{\textbf{DATE}} 
& \multicolumn{3}{c}{\textbf{LOC}} 
& \multicolumn{3}{c}{\textbf{ORG}} 
& \multicolumn{3}{c}{\textbf{PER}} 
& \multicolumn{3}{c}{\textbf{Overall}} \\
\cmidrule(lr){2-4}\cmidrule(lr){5-7}\cmidrule(lr){8-10}\cmidrule(lr){11-13}\cmidrule(lr){14-16}
\textbf{Model} 
& \textbf{P} & \textbf{R} & \textbf{F1} 
& \textbf{P} & \textbf{R} & \textbf{F1} 
& \textbf{P} & \textbf{R} & \textbf{F1} 
& \textbf{P} & \textbf{R} & \textbf{F1} 
& \textbf{P} & \textbf{R} & \textbf{F1} \\
\midrule

Afro-XLMR &
\textbf{0.760} & \textbf{0.790} & \textbf{0.770} &
0.870 & \textbf{0.890} & \textbf{0.880} &
0.740 & \textbf{0.720} & \textbf{0.730} &
0.850 & 0.820 & 0.830 &
0.818 & \textbf{0.833} & \textbf{0.826} \\

XLM-R &
0.750 & 0.710 & 0.730 &
0.870 & 0.880 & 0.870 &
\textbf{0.780} & 0.670 & 0.720 &
0.830 & 0.820 & 0.820 &
\textbf{0.823} & 0.798 & 0.810 \\

mBERT &
0.720 & 0.720 & 0.720 &
0.850 & 0.880 & 0.870 &
0.730 & 0.680 & 0.710 &
\textbf{0.890} & \textbf{0.840} & \textbf{0.860} &
0.804 & 0.803 & 0.803 \\

\bottomrule
\end{tabular}
}
\caption{Monolingual fine-tuning results on RunyaNER (Baseline). Precision (P), recall (R), and F$_1$-score (F1) are reported for each entity type and overall. }
\label{tab:mono_all_results}
\end{table*} 

\begin{table*}[t!]
\centering
\resizebox{\textwidth}{!}{
\begin{tabular}{l l l c c c c c c c c}
\toprule
\textbf{Code} & \textbf{Language} & \textbf{Subgroup}
& \multicolumn{4}{c}{\textbf{Zero-shot F$_1$}}
& \multicolumn{4}{c}{\textbf{Multilingual F$_1$}} \\
\cmidrule(lr){4-7} \cmidrule(lr){8-11}
& & & \textbf{Afro-XLMR} & \textbf{mBERT} & \textbf{XLM-R} & \textbf{Mean}
& \textbf{Afro-XLMR} & \textbf{mBERT} & \textbf{XLM-R} & \textbf{Mean} \\
\midrule
kin & Kinyarwanda     & Bantu (Great Lakes)  & 0.619 & 0.576 & 0.557 & 0.584 & 0.820 & 0.813 & 0.814 & 0.816 \\
lug & Luganda         & Bantu (Great Lakes)  & 0.622 & 0.521 & 0.495 & 0.546 & 0.831 & 0.817 & 0.806 & 0.818 \\
nya & Chichewa/Nyanja & Bantu (Southeast)    & 0.576 & 0.462 & 0.447 & 0.495 & 0.814 & 0.808 & 0.818 & 0.813 \\
swa & Kiswahili       & Bantu (Swahili)      & 0.584 & 0.500 & 0.341 & 0.475 & 0.827 & 0.812 & 0.810 & 0.816 \\
wol & Wolof           & Senegambian          & 0.464 & 0.481 & 0.450 & 0.465 & 0.821 & 0.807 & 0.813 & 0.814 \\
twi & Akan/Twi        & Kwa                  & 0.527 & 0.398 & 0.377 & 0.434 & 0.817 & 0.810 & 0.807 & 0.811 \\
zul & isiZulu         & Bantu (Southern)     & 0.515 & 0.415 & 0.371 & 0.434 & 0.814 & 0.815 & 0.809 & 0.813 \\
hau & Hausa           & Chadic               & 0.577 & 0.294 & 0.336 & 0.402 & 0.828 & 0.813 & 0.810 & 0.817 \\
luo & Luo             & Nilotic              & 0.447 & 0.323 & 0.414 & 0.395 & 0.818 & 0.817 & 0.810 & 0.815 \\
sna & chiShona        & Bantu (Southern)     & 0.483 & 0.307 & 0.352 & 0.381 & 0.824 & 0.816 & 0.814 & 0.818 \\
tsn & Setswana        & Bantu (Southern)     & 0.452 & 0.433 & 0.241 & 0.375 & 0.826 & 0.817 & 0.813 & 0.819 \\
xho & isiXhosa        & Bantu (Southern)     & 0.396 & 0.417 & 0.309 & 0.374 & 0.813 & 0.819 & 0.814 & 0.815 \\
ewe & Éwé             & Kwa                  & 0.389 & 0.263 & 0.441 & 0.364 & 0.823 & 0.804 & 0.793 & 0.807 \\
fon & Fon             & Volta-Niger          & 0.487 & 0.383 & 0.204 & 0.358 & 0.819 & 0.805 & 0.802 & 0.809 \\
bam & Bambara         & Mande                & 0.410 & 0.332 & 0.277 & 0.340 & 0.820 & 0.807 & 0.804 & 0.810 \\
ibo & Igbo            & Volta-Niger          & 0.481 & 0.243 & 0.269 & 0.331 & 0.828 & 0.813 & 0.809 & 0.817 \\
mos & Mossi (Mooré)   & Gur                  & 0.355 & 0.330 & 0.263 & 0.316 & 0.812 & 0.810 & 0.810 & 0.811 \\
yor & Yorùbá          & Volta-Niger          & 0.475 & 0.174 & 0.247 & 0.299 & 0.822 & 0.811 & 0.810 & 0.814 \\
pcm & Naijá Pidgin    & English-based        & 0.387 & 0.170 & 0.143 & 0.233 & 0.826 & 0.813 & 0.809 & 0.816 \\
bbj & Ghomálá'        & Grassfields          & 0.052 & 0.096 & 0.025 & 0.058 & 0.815 & 0.803 & 0.804 & 0.807 \\
\bottomrule
\end{tabular}
}
\caption{Zero-shot cross-lingual and multilingual (bilingual) fine-tuning F$_1$ results on the RunyaNER test set for all auxiliary languages, ordered by zero-shot mean F$_1$.}
\label{tab:combined_results}
\end{table*}

\section{Benchmarking PLMs on RunyaNER}
\label{sec:finetuning_plms}

This section establishes baseline results for Runyankore NER using multilingual PLMs under three supervision regimes: monolingual fine-tuning, cross-lingual zero-shot transfer, and multilingual (bilingual) fine-tuning with one auxiliary language. The aim is to compare how model pretraining and supervision regime affect Runyankore NER, and to provide a controlled empirical foundation for the auxiliary language selection analysis in Section~\ref{sec:aux_selection}.

\subsection{Experimental Setup}

We evaluate three encoder-only multilingual PLMs: mBERT \citep{devlin-etal-2019-bert}, XLM-R \citep{conneau-etal-2020-unsupervised}, and Afro-XLMR \citep{alabi-etal-2022-adapting}. 

Afro-XLMR adapts XLM-R through continued pretraining on 17 African languages, many of which are related to Runyankore.
This comparison allows us to examine whether broad multilingual coverage alone (mBERT, XLM-R) is sufficient for Runyankore, or whether regionally adapted pretraining yields additional gains.
We follow the MasakhaNER fine-tuning setup \citep{adelani-etal-2022-masakhaner}, training all our models for 10 epochs with AdamW \citep{loshchilov2019decoupledweightdecayregularization}, a maximum sequence length of 164, and a batch size of 32. %

For evaluation, we report span-level micro-averaged precision, recall, and F$_1$.

We consider three fine-tuning regimes. 
In the \textit{monolingual} setting, we fine-tune only on the RunyaNER training split and evaluate on its test set. In the \textit{zero-shot} setting, we fine-tune on MasakhaNER~2.0 training data for one auxiliary language and evaluate directly on the RunyaNER test set without exposure to Runyankore training data. In \textit{multilingual (bilingual) fine-tuning}, we fine-tune on a combination of RunyaNER training data and MasakhaNER~2.0 training data from one auxiliary language, still evaluating on the RunyaNER test set.
The RunyaNER development and test partitions are used only for NER evaluation; they are never used to train a model or to construct a similarity metric.
We experiment with all 20 languages in MasakhaNER~2.0 (listed in Table~\ref{tab:combined_results}) as auxiliary languages.

\subsection{Results}

Table~\ref{tab:mono_all_results} reports overall and entity-level results for monolingual fine-tuning. All three models achieve performance levels that reflect reasonably effective NER capabilities, with Afro-XLMR achieving the highest overall F$_1$, followed by XLM-R and then mBERT. 

This pattern holds across most entity types and is also reflected across precision and recall, with a small bias towards higher recall in Afro-XLMR. 

Afro-XLMR is considered a strong and well-established PLM for African languages. The fact that it achieves the best performance here reaffirms the value of regionally focussed multilingual PLMs. Even though Runyankore was not included in Afro-XLMR pretraining or adaptation, related Bantu languages such as Kinyarwanda and Kirundi were, which benefits RunyaNER through cross-lingual transfer.

Table~\ref{tab:combined_results} summarises Runyankore NER performance under cross-lingual zero-shot transfer and multilingual fine-tuning.  

For zero-shot transfer, performance varies greatly across auxiliary languages. 
The strongest transfer consistently comes from Great Lakes Bantu languages (Luganda and Kinyarwanda), which are geographically and genealogically related to Runyankore. 

Performance declines for less related languages. 

This indicates that zero-shot transfer is highly sensitive to auxiliary language similarity. Afro-XLMR is generally more robust than mBERT and XLM-R in this regime, suggesting that regionally adapted pretraining is beneficial when the model must rely entirely on cross-lingual generalisation.

In the multilingual (bilingual) fine-tuning setting, adding one auxiliary language yields only modest improvements over the Runyankore monolingual baseline. The gains are largest for mBERT, whereas Afro-XLMR and XLM-R remain close to their monolingual results. This pattern suggests that models with broader multilingual coverage benefit more from additional cross-lingual supervision, while regionally adapted models already capture much of the transferable signal needed for Runyankore NER. The highest-ranked auxiliaries are again predominantly Bantu languages, although some more distant languages are also competitive once Runyankore supervision is included.

These results show that pretraining choice matters most under weak supervision. When Runyankore training data is available, all three models perform strongly and gaps are relatively small. In cross-lingual zero-shot transfer, however, the choice of model and auxiliary language becomes critical, with Afro-XLMR and closely related Bantu languages providing the strongest results.

\section{Auxiliary Language Selection for Cross-Lingual Transfer}
\label{sec:aux_selection}

The substantial variation in performance across auxiliary languages motivates examining whether useful transfer languages can be identified systematically using metadata-based, typological, or embedding-based similarity signals.
In this section, we investigate which measure of language similarity is most predictive for selecting auxiliary languages that optimise transfer for Runyankore NER. Metadata and typological measures require no Runyankore NER labels; the embedding-based measures use gold spans from the training split only.

\begin{table}[t]
\centering
\footnotesize
\setlength{\tabcolsep}{2pt}
\begin{tabular}{l cc cc cc}
\toprule
& \multicolumn{2}{c}{\textbf{mBERT}} & \multicolumn{2}{c}{\textbf{XLM-R}} & \multicolumn{2}{c}{\textbf{Afro-XLMR}} \\
\cmidrule(lr){2-3}\cmidrule(lr){4-5}\cmidrule(lr){6-7}
\textbf{Metric} & \textbf{0-shot} & \textbf{multi} & \textbf{0-shot} & \textbf{multi} & \textbf{0-shot} & \textbf{multi} \\
\midrule
LinguaMeta & 0.24 & 0.34 & 0.34 & 0.18 & \underline{\textbf{0.56}}$^{*}$ & \textbf{0.25} \\
URIEL      & \textbf{0.53}$^{*}$ & \textbf{0.51}$^{*}$ & \textbf{0.41} & \textbf{0.51}$^{*}$ & 0.52$^{*}$ & 0.10 \\
\midrule
Cosine     & 0.595$^{**}$ & \underline{\textbf{0.571}}$^{**}$ & \underline{\textbf{0.638}}$^{**}$ & \underline{\textbf{0.634}}$^{**}$ & \textbf{0.488}$^{*}$ & 0.239 \\
SWD        & \underline{\textbf{0.635}}$^{**}$ & 0.561$^{*}$ & 0.635$^{**}$ & 0.337 & 0.466$^{*}$ & \underline{\textbf{0.340}} \\
\bottomrule
\end{tabular}
\caption{Spearman correlation ($\rho$) between similarity metrics and Runyankore NER $F_1$ across $n=20$ auxiliary languages. Embedding-based $\rho$ is the strongest layer. We \textbf{boldface} the highest $\rho$ per metric type and \underline{\textbf{underline}} the highest $\rho$ overall. $^{*}$~$p<0.05$, $^{**}$~$p<0.01$ (two-tailed).}
\label{tab:similarity_correlation_summary}
\end{table}

\subsection{Language Similarity Metrics}

We compare four methods, two of which rely on traditional linguistic resources (metadata and typological features), and two of which are based on model representation (embedding) similarity (cosine similarity and Sliced Wasserstein Distance). We evaluate their ability to predict downstream performance under two regimes: cross-lingual zero-shot transfer and multilingual (bilingual) fine-tuning. Transfer effectiveness is measured using entity-level span $F_1$, and correlations are computed across 20 auxiliary languages and three encoders (mBERT, XLM-R, Afro-XLMR).

The four metrics differ in the target-language supervision they require. LinguaMeta and URIEL are computed from published metadata and typological inventories and need no Runyankore NER labels. Prototype cosine similarity and SWD compare gold entity-span embeddings. For both embedding metrics, we extract spans only from the RunyaNER training split and from the MasakhaNER~2.0 training splits of the auxiliary languages. The development and test partitions are never used to construct prototypes or span distributions. Embedding-based selection is therefore not a no-label procedure: it assumes a labelled target training set, or an equivalent seed of gold spans. What it does not use is the evaluation set, so the subsequent transfer experiments remain uncontaminated. We treat LinguaMeta and URIEL as the label-free baselines against which these span-based measures are compared.

\paragraph{Metadata similarity (LinguaMeta).}
This method uses structured metadata such as script, geographic region, and language location to estimate cross-lingual relatedness \citep{ritchie-etal-2024-linguameta-unified}. Individual features are normalised and combined into a single similarity score, capturing coarse-grained signals such as script compatibility and regional overlap. While simple and interpretable, this approach primarily reflects geographic and sociolinguistic proximity rather than linguistic structure.

\paragraph{Typological similarity (URIEL/lang2vec).}
This method represents languages using typological feature vectors derived from URIEL \citep{littell-etal-2017-uriel}, accessed via lang2vec. We use concatenated syntactic and phonological vectors derived from resources such as WALS \citep{wals} and PHOIBLE \citep{phoible}. Similarity is computed using cosine similarity between these vectors, yielding a linguistically grounded measure that reflects structural similarity between languages.

\paragraph{Prototype cosine similarity.}
This representation-based metric operates directly in multilingual embedding space. For each language, gold entity spans from the corresponding training split are encoded using a pretrained multilingual model and aggregated into entity-type prototypes via mean pooling. Similarity is computed as cosine similarity between corresponding Runyankore and auxiliary prototypes, capturing alignment of named entities in the model's representation space. This method is layer-dependent, and similarity is computed separately at each encoder layer. Recent work has shown that model-based similarity measures can outperform hand-crafted feature-based methods for transfer prediction \citep{deshpande2022when, dou2021word, muller2021first, ebrahimi-etal-2025-model}.

\begin{figure*}[t!]
\centering
\includegraphics[width=\textwidth]{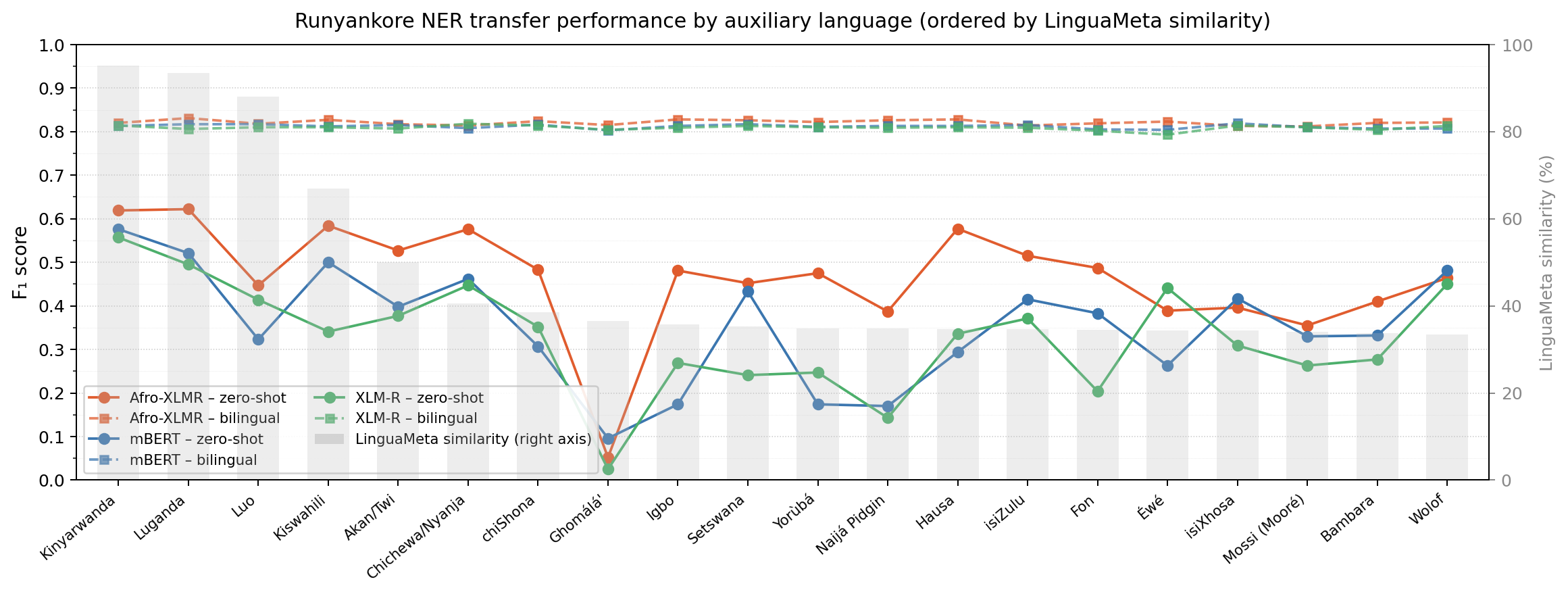}
\caption{Runyankore NER transfer performance by auxiliary language ordered according to LinguaMeta similarity relative to Runyankore. Languages are arranged from highest to lowest similarity. Solid lines indicate cross-lingual zero-shot transfer, while dashed lines indicate multilingual fine-tuning performance.}
\label{fig:linguameta_transfer}
\end{figure*}

\begin{figure*}[t!]
\centering
\includegraphics[width=\textwidth]{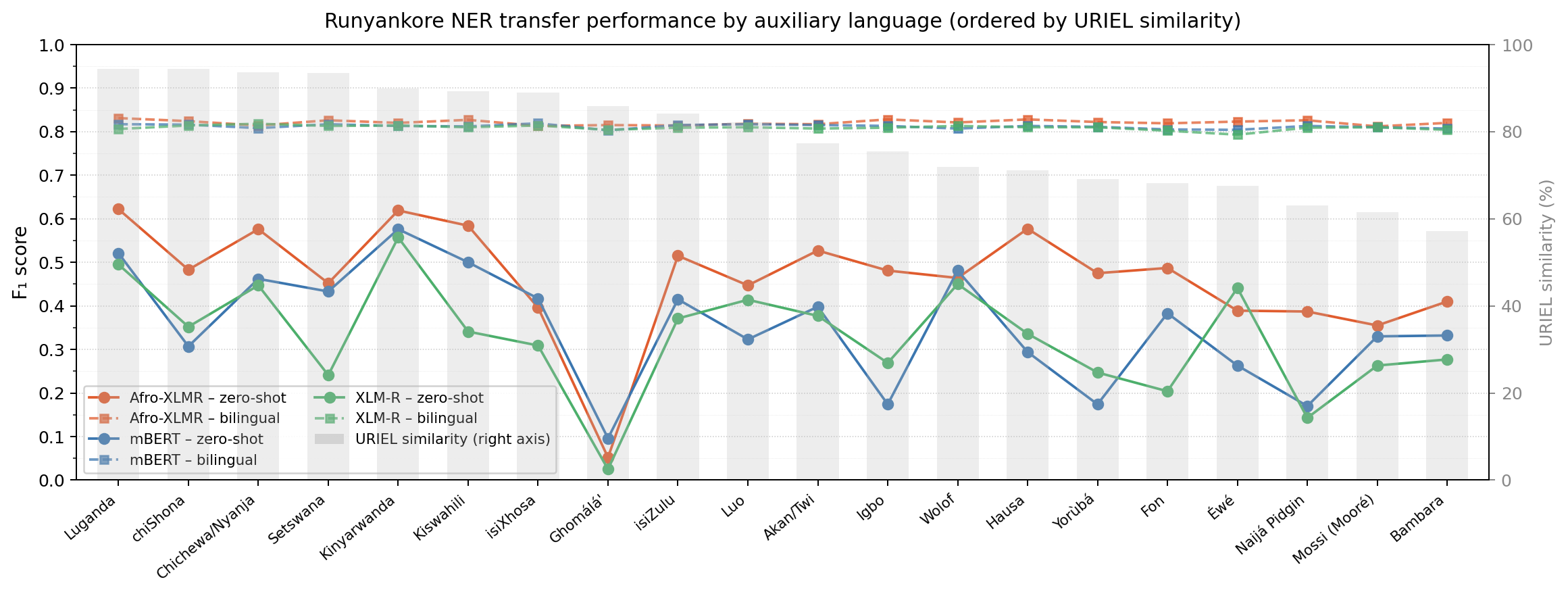}
\caption{Runyankore NER transfer performance by auxiliary language ordered according to URIEL typological similarity relative to Runyankore.}
\label{fig:uriel_transfer}
\end{figure*}

\paragraph{Sliced Wasserstein Distance (SWD).}
SWD compares full distributions of entity-span embeddings rather than only their means, using the same training-split spans as the cosine prototypes. It approximates the Wasserstein-1 distance by projecting embeddings onto multiple random directions and averaging one-dimensional distances \citep{nguyen2020distributionalslicedwassersteinapplicationsgenerative}. Unlike cosine similarity, SWD captures higher-order geometric differences such as dispersion and multimodality, providing a richer notion of representational similarity.

\subsection{Which similarity metrics predict transfer?}

We evaluate predictive usefulness by computing Spearman $\rho$ between similarity scores and Runyankore NER $F_1$ across $n=20$ auxiliary languages. For LinguaMeta and URIEL, we use a single similarity score per language. For prototype cosine similarity and SWD, which are layer-dependent, we report the strongest observed Spearman correlation across layers. Because this best-layer choice is optimistic, we treat those coefficients as an upper bound on the layer-wise signal rather than as a pre-specified test. Table~\ref{tab:similarity_correlation_summary} marks two-tailed significance under the standard $t$ approximation for Spearman $\rho$ with $n=20$ (critical $|\rho|\approx 0.44$ at $p<0.05$ and $|\rho|\approx 0.56$ at $p<0.01$).

Correlation coefficients are presented in Table~\ref{tab:similarity_correlation_summary}. Coefficients are higher for zero-shot transfer than multilingual fine-tuning, confirming that auxiliary language selection is most critical when no Runyankore supervision is available for the NER model itself.
Among linguistically-informed methods, typological similarity (URIEL) provides a more consistent signal than language metadata (LinguaMeta), although several LinguaMeta coefficients are not significant at $n=20$.
Embedding-based metrics are the most informative: the strongest cosine and SWD zero-shot correlations reach $p<0.01$, but the Afro-XLMR multilingual cosine coefficient and several SWD multilingual coefficients are not significant.
These results support embedding-based similarity as a stronger predictor of transfer than metadata or typology, while remaining compatible with a modest sample of twenty auxiliaries.

\begin{figure*}[t]
\centering
\includegraphics[width=\textwidth]{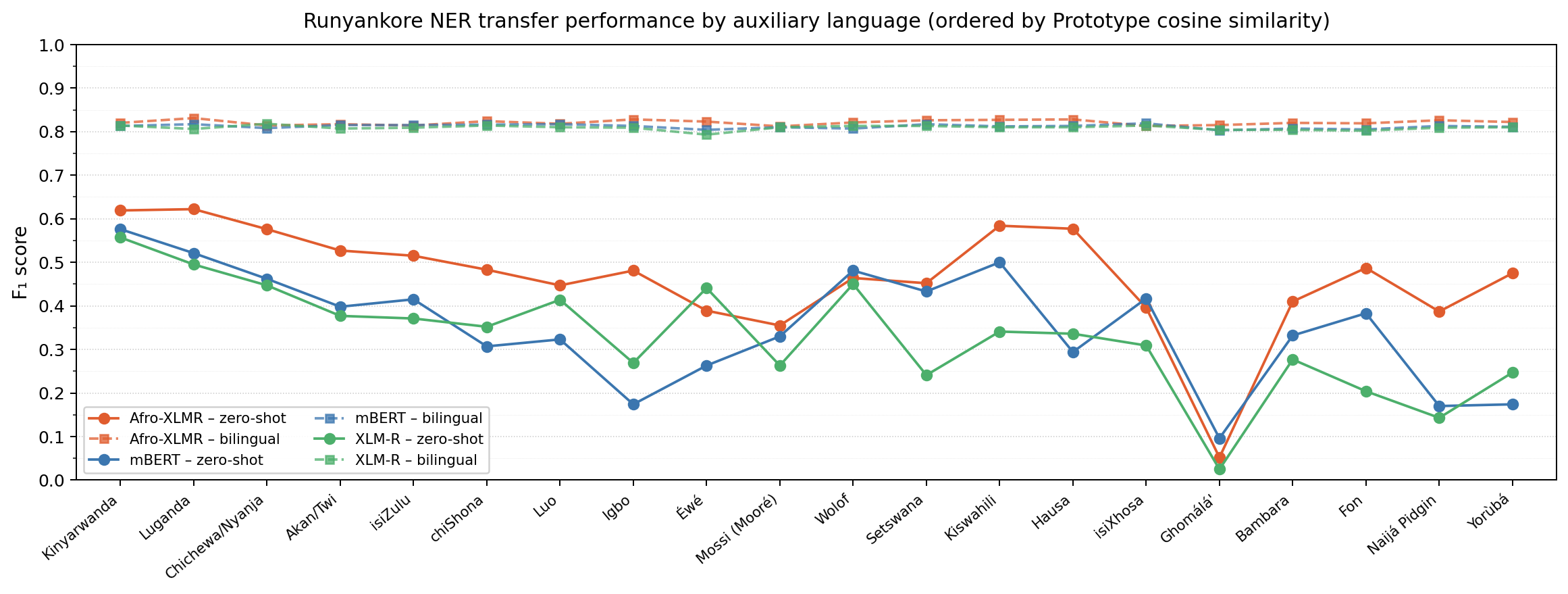}
\caption{Runyankore NER transfer performance by auxiliary language ordered according to prototype cosine similarity relative to Runyankore.}
\label{fig:cosine_transfer}
\end{figure*}

\begin{figure*}[t]
\centering
\includegraphics[width=\textwidth]{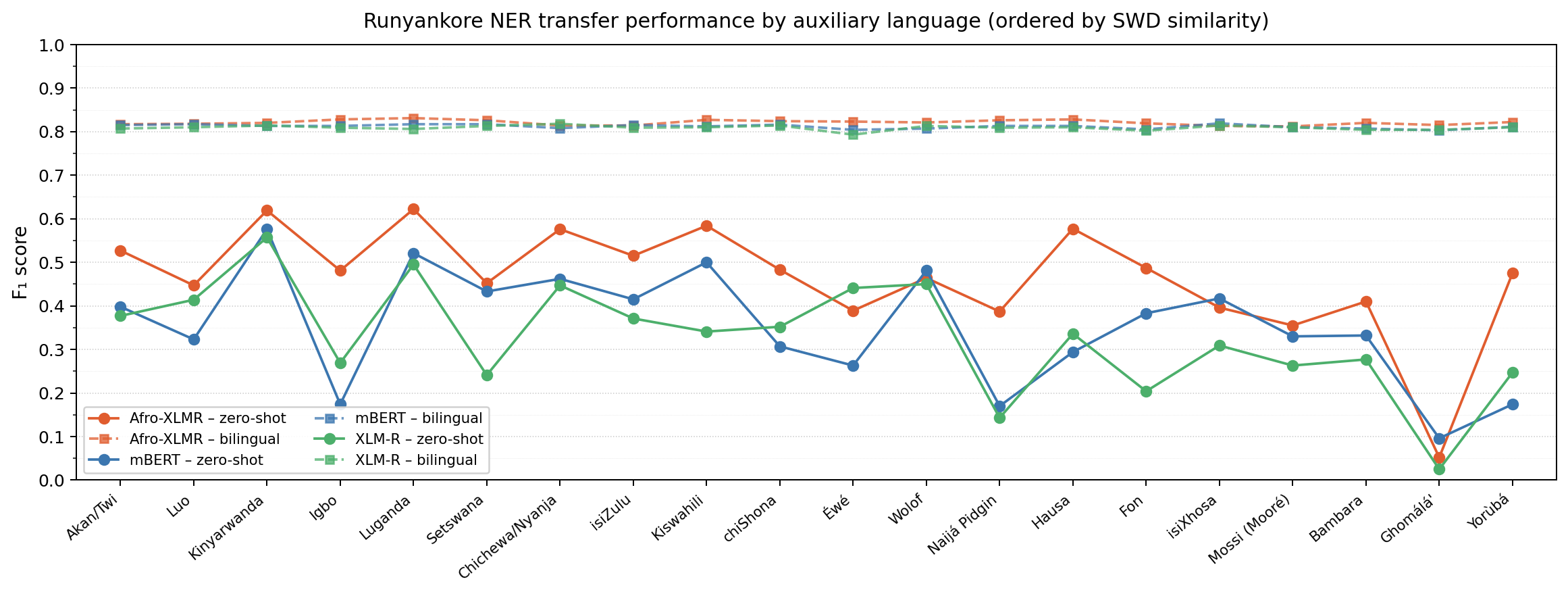}
\caption{Runyankore NER transfer performance by auxiliary language ordered according to Sliced Wasserstein Distance (SWD) similarity relative to Runyankore. Lower SWD values indicate stronger representational similarity.}
\label{fig:swd_transfer}
\end{figure*} 

\subsection{Similarity ordering and transfer performance}
\label{ss:similarity-results} 

Figures~\ref{fig:linguameta_transfer}, \ref{fig:uriel_transfer}, \ref{fig:cosine_transfer}, and \ref{fig:swd_transfer} plot Runyankore NER transfer performance across auxiliary languages, ordered according to each similarity metric. 

These plots provide a detailed visualisation of the results underlying the trends in Table~\ref{tab:similarity_correlation_summary}. The weaker correlations of linguistic resource-based methods are reflected in the more unstable performance patterns of Figures~\ref{fig:linguameta_transfer} and \ref{fig:uriel_transfer}. While these metrics are broadly predictive (very high-similarity languages tend to transfer better than very low-similarity ones), performance varies substantially across individual language rankings.

In contrast, embedding-based similarity measures produce more consistent alignment between similarity ordering and transfer effectiveness, as shown in Figure \ref{fig:cosine_transfer}, and \ref{fig:swd_transfer}. This is especially visible for prototype cosine similarity in Figure \ref{fig:cosine_transfer}, where we see a clear relationship between similarity and transfer performance for the languages closest to Runyankore in multilingual embedding space. Embedding-based cosine similarity offers a more precise signal than linguistic resource-based metrics, distinguishing between individual top-ranked auxiliary language candidates rather than only separating highly dissimilar languages.

The plots also show a strong regime-dependent pattern. In the cross-lingual zero-shot setting, transfer performance varies substantially across auxiliary languages, indicating that auxiliary language selection is important when no Runyankore supervision is available. In contrast, multilingual fine-tuning produces substantially flatter transfer curves, suggesting that auxiliary language similarity becomes less important once modest amounts of Runyankore supervision are introduced.

\begin{table*}[t]
\centering
\resizebox{\textwidth}{!}{
\begin{tabular}{l l c c c c c c c c}
\toprule
\textbf{Similarity} & \textbf{Auxiliary}
& \multicolumn{4}{c}{\textbf{Zero-shot F$_1$}}
& \multicolumn{4}{c}{\textbf{Multilingual F$_1$}} \\
\cmidrule(lr){3-6} \cmidrule(lr){7-10}
\textbf{metric} & \textbf{language(s)} & \textbf{Afro-XLMR} & \textbf{mBERT} & \textbf{XLM-R} & \textbf{Mean}
& \textbf{Afro-XLMR} & \textbf{mBERT} & \textbf{XLM-R} & \textbf{Mean} \\
\midrule

\multicolumn{10}{l}{\textit{Quartets}} \\
LinguaMeta & lug/kin/swa/luo
& 0.644 & 0.615 & 0.635 & 0.631
& 0.817 & \textbf{0.821} & 0.809 & 0.816 \\

URIEL & tsn/nya/lug/swa
& 0.593 & 0.582 & 0.566 & 0.580
& 0.812 & 0.813 & \textbf{0.816} & 0.814 \\

Cosine & lug/sna/kin/nya
& \textbf{0.651} & \textbf{0.619} & \textbf{0.653} & \textbf{0.641}
& 0.825 & 0.815 & 0.814 & \textbf{0.818} \\

SWD & kin/twi/luo/ibo
& 0.598 & 0.515 & 0.562 & 0.558
& 0.817 & 0.815 & 0.809 & 0.814 \\

\midrule

\multicolumn{10}{l}{\textit{Single auxiliary}} \\

LinguaMeta & kin
& 0.619 & 0.576 & 0.557 & 0.584
& 0.820 & 0.813 & 0.814 & 0.816 \\

URIEL & lug
& 0.622 & 0.521 & 0.495 & 0.546
& \textbf{0.831} & 0.817 & 0.806 & \textbf{0.818} \\

Cosine & kin
& 0.619 & 0.576 & 0.557 & 0.584
& 0.817 & 0.820 & 0.814 & 0.817 \\

SWD & twi
& 0.527 & 0.398 & 0.377 & 0.434
& 0.817 & 0.810 & 0.807 & 0.811 \\

\bottomrule
\end{tabular}
}
\caption{Runyankore NER performance under different auxiliary language selection schemes. Top-ranked quartets are shown for each similarity scheme, alongside the most similar single auxiliary language per scheme. \textbf{Bold} indicates the highest value in each column.}
\label{tab:quartet_transfer_full}
\end{table*}

\subsection{From individual languages to auxiliary language groups}

In many multilingual transfer settings, supervision is drawn from groups of auxiliary languages rather than a single source language \citep{shaffer-2021-language-clustering, ebrahimi-etal-2025-model}. We therefore extend the analysis from individual auxiliary languages to auxiliary language quartets. For each similarity metric, languages were first ranked according to their similarity to Runyankore. We then constructed language quartets by selecting the four highest-ranked auxiliary languages under each similarity scheme (the first two columns of Table~\ref{tab:quartet_transfer_full} list the resulting language groups). For layer-dependent embedding-based metrics, rankings were averaged across all layers and the three multilingual PLMs to obtain a model-agnostic ordering. The resulting quartets therefore represent the strongest auxiliary language groups predicted by each similarity metric.

Table~\ref{tab:quartet_transfer_full} reports F$_1$ scores for the highest-ranked quartet under each method. 
We compare these results to the top-ranked single auxiliary language predicted by each similarity metric (similarly averaged for the embedding-based metrics).
In the zero-shot setting, using multiple auxiliary languages consistently outperforms using a single auxiliary language, with quartets improving mean F$_1$ by several points. In multilingual fine-tuning, this advantage largely disappears: differences between quartets and single auxiliaries are small, suggesting that once Runyankore supervision is available, additional auxiliary languages offer little benefit.

The results reaffirm that, in cross-lingual zero-shot transfer, auxiliary language selection is crucial. Cosine-based quartets consistently achieve the best performance across all models, followed by LinguaMeta and SWD. As in individual auxiliary language selection, embedding representation-based similarity is more effective for selecting transfer languages.

Performance differences between quartets are much smaller in multilingual fine-tuning than in zero-shot transfer. Performance converges across grouping strategies, and no single method consistently dominates. As in individual auxiliary language selection, once modest Runyankore supervision is available, the importance of auxiliary group language selection is substantially reduced. 

These results reinforce the earlier findings. Auxiliary language selection is critical under cross-lingual zero-shot transfer, where embedding-based similarity provides the strongest guidance. Under multilingual fine-tuning, however, the effect of auxiliary choice is substantially reduced. 


\section{Conclusion}
\label{sec:conclusion}

We introduced RunyaNER, the first publicly available NER dataset for Runyankore, and established benchmark results for Runyankore NER using multilingual PLMs under monolingual, zero-shot, and multilingual fine-tuning settings. 
Our experiments show that auxiliary language selection is particularly important in cross-lingual zero-shot transfer, where closely related Great Lakes Bantu languages yield the strongest transfer performance. Embedding-based similarity measures, computed from labelled training spans rather than from the evaluation set, correlate more strongly with downstream transfer effectiveness than metadata-based or typological approaches, while auxiliary language choice becomes substantially less important once Runyankore supervision is introduced through multilingual fine-tuning.
We hope that RunyaNER will support future research on low-resource African NLP and multilingual transfer learning.


\section*{Limitations}

This study focuses only on Runyankore NER, and the observed auxiliary-language selection patterns may not generalise to other NLP tasks or languages. Our experiments are limited to three encoder-based multilingual models and twenty African auxiliary languages, which restricts broader architectural and linguistic coverage and yields modest power for Spearman tests ($n=20$). RunyaNER was annotated and verified by a single L1 speaker, so we cannot report inter-annotator agreement; quality rests on exhaustive guideline-based correction rather than multi-annotator adjudication. In addition, the two sources of the dataset are concentrated in relatively similar domains. Embedding-based similarity measures require gold entity spans from the target training split and are therefore not a label-free selection method; they also depend on modelling choices such as encoder layer selection and embedding aggregation, which were not evaluated exhaustively. We report the strongest layer-wise correlation for those metrics, which should be read as an upper bound.

\section*{Acknowledgements}

Computations were performed using facilities provided by the University of Cape Town’s ICTS High Performance Computing team: \url{hpc.uct.ac.za}.

\bibliography{custom}

\clearpage
\appendix
\section*{Appendix}

\renewcommand{\thetable}{A.\arabic{table}}
\setcounter{table}{0}
 
\section{Embedding-Based Similarity Results}

\begin{table}[H]
\centering
\vspace{1em}
\tiny
\setlength{\tabcolsep}{1.6pt}
\renewcommand{\arraystretch}{0.78}

\begin{minipage}{\textwidth}
\centering
\resizebox{\textwidth}{!}{
\begin{tabular}{S[table-format=2.2] l l l
                l l
                l l
                l l}
\toprule
{\textbf{Avg.\ Rank}} & \textbf{Code} & \textbf{Language} & \textbf{Subgroup}
& \multicolumn{2}{c}{\textbf{Afro-XLMR}}
& \multicolumn{2}{c}{\textbf{mBERT}}
& \multicolumn{2}{c}{\textbf{XLM-R}} \\
\cmidrule(lr){5-6} \cmidrule(lr){7-8} \cmidrule(lr){9-10}
& & & & {Mean $\pm$ Std} & {Rank} & {Mean $\pm$ Std} & {Rank} & {Mean $\pm$ Std} & {Rank} \\
\midrule
1.33 & kin & Kinyarwanda     & Bantu (Great Lakes)
& 0.9984$\pm$0.0040 & (2) & 0.9913$\pm$0.0069 & (1) & 0.999989$\pm$0.000019 & (1) \\
3.33 & lug & Luganda         & Bantu (Great Lakes)
& 0.9979$\pm$0.0055 & (4) & 0.9899$\pm$0.0084 & (4) & 0.999986$\pm$0.000030 & (2) \\
4.33 & nya & Chichewa/Nyanja & Bantu (Southeast)
& 0.9976$\pm$0.0065 & (7) & 0.9904$\pm$0.0077 & (3) & 0.999985$\pm$0.000029 & (3) \\
4.67 & luo & Luo             & Nilotic
& 0.9981$\pm$0.0047 & (3) & 0.9880$\pm$0.0096 & (7) & 0.999982$\pm$0.000029 & (4) \\
5.00 & twi & Akan/Twi        & Kwa
& 0.9985$\pm$0.0033 & (1) & 0.9891$\pm$0.0086 & (5) & 0.999972$\pm$0.000057 & (9) \\
7.00 & zul & isiZulu         & Bantu (Southern)
& 0.9975$\pm$0.0063 & (9) & 0.9884$\pm$0.0093 & (6) & 0.999979$\pm$0.000045 & (6) \\
7.33 & sna & chiShona        & Bantu (Southern)
& 0.9970$\pm$0.0081 & (12) & 0.9910$\pm$0.0071 & (2) & 0.999974$\pm$0.000072 & (8) \\
8.67 & ibo & Igbo            & Volta-Niger
& 0.9976$\pm$0.0065 & (8) & 0.9879$\pm$0.0095 & (8) & 0.999972$\pm$0.000054 & (10) \\
9.67 & ewe & Éwé             & Kwa
& 0.9977$\pm$0.0053 & (6) & 0.9815$\pm$0.0151 & (16) & 0.999975$\pm$0.000046 & (7) \\
10.67 & mos & Mossi (Mooré)  & Gur
& 0.9972$\pm$0.0073 & (10) & 0.9865$\pm$0.0108 & (11) & 0.999971$\pm$0.000054 & (11) \\
10.67 & wol & Wolof          & Senegambian
& 0.9969$\pm$0.0079 & (15) & 0.9849$\pm$0.0118 & (12) & 0.999979$\pm$0.000037 & (5) \\
11.33 & tsn & Setswana       & Bantu (Southern)
& 0.9977$\pm$0.0053 & (5) & 0.9825$\pm$0.0141 & (14) & 0.999967$\pm$0.000089 & (15) \\
12.33 & swa & Kiswahili      & Bantu (Swahili)
& 0.9970$\pm$0.0059 & (11) & 0.9840$\pm$0.0135 & (13) & 0.999970$\pm$0.000053 & (13) \\
14.33 & hau & Hausa          & Chadic
& 0.9967$\pm$0.0071 & (16) & 0.9878$\pm$0.0097 & (9) & 0.999910$\pm$0.000257 & (18) \\
14.67 & xho & isiXhosa       & Bantu (Southern)
& 0.9963$\pm$0.0098 & (20) & 0.9877$\pm$0.0099 & (10) & 0.999967$\pm$0.000082 & (14) \\
16.00 & bbj & Ghomálá’       & Grassfields
& 0.9966$\pm$0.0088 & (17) & 0.9742$\pm$0.0220 & (19) & 0.999971$\pm$0.000061 & (12) \\
16.67 & bam & Bambara        & Mande
& 0.9966$\pm$0.0093 & (18) & 0.9821$\pm$0.0155 & (15) & 0.999956$\pm$0.000086 & (17) \\
16.67 & fon & Fon            & Volta-Niger
& 0.9970$\pm$0.0072 & (14) & 0.9766$\pm$0.0190 & (17) & 0.999892$\pm$0.000277 & (19) \\
17.67 & pcm & Naijá Pidgin   & English-based
& 0.9970$\pm$0.0055 & (13) & 0.9726$\pm$0.0233 & (20) & 0.999828$\pm$0.000551 & (20) \\
17.67 & yor & Yorùbá         & Volta-Niger
& 0.9965$\pm$0.0093 & (19) & 0.9755$\pm$0.0185 & (18) & 0.999964$\pm$0.000064 & (16) \\
\bottomrule
\end{tabular}
}
\caption{Prototype cosine similarity between Runyankore and each auxiliary language, averaged across the 12 layers (mean $\pm$ std). Languages are ordered by average rank across Afro-XLMR, mBERT, and XLM-R (lower is better). Each model reports mean cosine similarity ($\pm$ std) and the within-model rank.}
\label{tab:cosine_merged_overall}
\end{minipage}

\vspace{5em}

\begin{minipage}{\textwidth}
\centering
\resizebox{\textwidth}{!}{
\begin{tabular}{S[table-format=2.2] l l l
                l l
                l l
                l l}
\toprule
{\textbf{Avg.\ Rank}} & \textbf{Code} & \textbf{Language} & \textbf{Subgroup}
& \multicolumn{2}{c}{\textbf{Afro-XLMR}}
& \multicolumn{2}{c}{\textbf{mBERT}}
& \multicolumn{2}{c}{\textbf{XLM-R}} \\
\cmidrule(lr){5-6} \cmidrule(lr){7-8} \cmidrule(lr){9-10}
& & & & {Mean $\pm$ Std} & {Rank} & {Mean $\pm$ Std} & {Rank} & {Mean $\pm$ Std} & {Rank} \\
\midrule
1.33 & twi & Akan/Twi & Kwa
& 0.0058$\pm$0.0012 & (1) & 0.0092$\pm$0.0018 & (1) & 0.0056$\pm$0.0015 & (2) \\
1.67 & kin & Kinyarwanda & Bantu (Great Lakes)
& 0.0059$\pm$0.0010 & (2) & 0.0095$\pm$0.0018 & (2) & 0.0054$\pm$0.0014 & (1) \\
3.00 & luo & Luo & Nilotic
& 0.0061$\pm$0.0011 & (3) & 0.0097$\pm$0.0020 & (3) & 0.0060$\pm$0.0016 & (3) \\
5.33 & ibo & Igbo & Volta-Niger
& 0.0064$\pm$0.0011 & (7) & 0.0103$\pm$0.0018 & (5) & 0.0061$\pm$0.0015 & (4) \\
6.33 & lug & Luganda & Bantu (Great Lakes)
& 0.0062$\pm$0.0011 & (4) & 0.0100$\pm$0.0018 & (4) & 0.0068$\pm$0.0017 & (11) \\
6.67 & tsn & Setswana & Bantu (Southern)
& 0.0064$\pm$0.0011 & (6) & 0.0106$\pm$0.0019 & (9) & 0.0063$\pm$0.0015 & (5) \\
8.33 & nya & Chichewa/Nyanja & Bantu (Southeast)
& 0.0065$\pm$0.0011 & (8) & 0.0104$\pm$0.0019 & (7) & 0.0067$\pm$0.0017 & (10) \\
8.33 & zul & isiZulu & Bantu (Southern)
& 0.0066$\pm$0.0011 & (9) & 0.0109$\pm$0.0022 & (10) & 0.0064$\pm$0.0017 & (6) \\
8.67 & ewe & Éwé & Kwa
& 0.0063$\pm$0.0011 & (5) & 0.0117$\pm$0.0022 & (14) & 0.0065$\pm$0.0018 & (7) \\
9.67 & swa & Kiswahili & Bantu (Swahili)
& 0.0067$\pm$0.0011 & (10) & 0.0103$\pm$0.0018 & (6) & 0.0069$\pm$0.0018 & (13) \\
10.00 & sna & chiShona & Bantu (Southern)
& 0.0070$\pm$0.0013 & (13) & 0.0106$\pm$0.0021 & (8) & 0.0066$\pm$0.0017 & (9) \\
10.33 & wol & Wolof & Senegambian
& 0.0069$\pm$0.0013 & (12) & 0.0113$\pm$0.0021 & (11) & 0.0066$\pm$0.0017 & (8) \\
13.33 & pcm & Naijá Pidgin & English-based
& 0.0068$\pm$0.0012 & (11) & 0.0114$\pm$0.0020 & (13) & 0.0072$\pm$0.0019 & (16) \\
15.33 & hau & Hausa & Chadic
& 0.0071$\pm$0.0013 & (14) & 0.0114$\pm$0.0022 & (12) & 0.0078$\pm$0.0021 & (20) \\
15.33 & xho & isiXhosa & Bantu (Southern)
& 0.0079$\pm$0.0014 & (19) & 0.0120$\pm$0.0021 & (15) & 0.0068$\pm$0.0018 & (12) \\
16.33 & fon & Fon & Volta-Niger
& 0.0073$\pm$0.0013 & (15) & 0.0122$\pm$0.0024 & (16) & 0.0073$\pm$0.0021 & (18) \\
16.33 & mos & Mossi (Mooré) & Gur
& 0.0077$\pm$0.0014 & (18) & 0.0124$\pm$0.0025 & (17) & 0.0071$\pm$0.0021 & (14) \\
16.67 & bbj & Ghomálá’ & Grassfields
& 0.0074$\pm$0.0014 & (16) & 0.0132$\pm$0.0026 & (19) & 0.0071$\pm$0.0019 & (15) \\
18.00 & bam & Bambara & Mande
& 0.0076$\pm$0.0014 & (17) & 0.0125$\pm$0.0022 & (18) & 0.0075$\pm$0.0020 & (19) \\
19.00 & yor & Yorùbá & Volta-Niger
& 0.0081$\pm$0.0015 & (20) & 0.0146$\pm$0.0028 & (20) & 0.0073$\pm$0.0020 & (17) \\
\bottomrule
\end{tabular}
}
\caption{Sliced Wasserstein Distance (SWD) results between Runyankore and each auxiliary language, averaged across the 12 layers (mean $\pm$ std). Languages are ordered by average rank across Afro-XLMR, mBERT, and XLM-R (lower is better). Each model reports mean SWD ($\pm$ std) and the within-model rank.}
\label{tab:swd_merged_overall}
\end{minipage}

\end{table}

\end{document}